\pdfoutput=1
\documentclass[11pt]{article}
\PassOptionsToPackage{dvipsnames}{xcolor}
\usepackage[final]{coling}

\usepackage{times}
\usepackage{latexsym}
\usepackage[bottom]{footmisc}
\usepackage[T1]{fontenc}
\usepackage[utf8]{inputenc}
\usepackage{microtype}
\usepackage{inconsolata}
\usepackage{graphicx}

\usepackage{graphicx}
\usepackage{amsmath}
\usepackage{bbold}
\DeclareMathOperator*{\argmax}{arg\,max}
\usepackage{booktabs}
\usepackage{pgfplotstable}
\usepackage{pgfplots}
\usepgfplotslibrary{groupplots}
\pgfplotsset{compat=1.9}
\usepackage{tikz}
\usepackage{caption}
\usepackage{subcaption}
\usepackage{multirow}
\usepackage{makecell}
\usepackage{float}
\usetikzlibrary{patterns}

\title{How Ambiguous Are the Rationales for Natural Language Reasoning? \\ A Simple Approach to Handling Rationale Uncertainty}

\author{Hazel Kim$^*$\\
  University of Oxford \\
  Classting AI Research \\
  \texttt{hazel.kim@cs.ox.ac.uk} 
}

\begin{document}
\maketitle
\def\thefootnote{*}\footnotetext{The work was mostly done at Classting AI Research. The author is currently affiliated with the University of Oxford.}\def\thefootnote{\arabic{footnote}}
\begin{abstract}
The quality of rationales is essential in the reasoning capabilities of language models. Rationales not only enhance reasoning performance in complex natural language tasks but also justify model decisions. However, obtaining impeccable rationales is often impossible. Our study aims to investigate how ambiguous rationales play in model performances of natural language reasoning. We first assess the ambiguity of rationales through the lens of entropy and uncertainty in model prior beliefs, exploring its impact on task performance. We then propose a simple way to guide models to choose between two different reasoning paths depending on the ambiguity of the rationales. Our empirical results demonstrate that this approach leads to robust performance, particularly in adversarial scenarios where rationale quality is inconsistent.
\end{abstract}

\section{Introduction}

\usepgfplotslibrary{units}
\begin{figure}[t!]
\centering
\pgfplotsset{
width=0.25\textwidth,
my style/.style={
            every axis plot/.append style={line width=0.8pt},
            font=\tiny,           
            symbolic x coords={1,2,3,4,5,6,7,8,9,10},
            xtick=data,
            xmin=1,xmax=10,
            ymin=30, ymax=110,
            grid,
            grid style={dotted},},
my legend style/.style={
            font=\scriptsize,
            legend entries={
                Train data,
                Validation data,     
            },
            legend style={
                draw=none,
                at={([yshift=2pt]1.2,1.1)},
                anchor=south,
            },
            legend columns=2,
        },
        every axis title/.append style={at={(0.5,-0.55)}},
        every axis x label/.style={at={(0.3,0.01)},anchor=south west},
        every axis y label/.style={at={(-0.15,0.0)},anchor=south west, rotate=90}
        }
\begin{tikzpicture}
\centering

\begin{groupplot}[
    group style={group size=2 by 1, 
                horizontal sep = 1cm,
                vertical sep = 1.2cm}]
    \nextgroupplot[my style, 
                  my legend style, 
                 y unit={\%},
                 xlabel=\# Epochs,
                 ylabel=Accuracy, 
                 title={\footnotesize{(a) CSQA}}]
        \addplot[color=BurntOrange, mark=none] table{csqa_tr.dat}; 
        \addplot[color=JungleGreen, mark=none] table{csqa_te.dat}; 
    \nextgroupplot[cycle list shift=1, 
                   my style, 
                   title={\footnotesize{(b) QASC }}]
        \addplot[color=BurntOrange, mark=none] table{qasc_tr.dat}; 
        \addplot[color=JungleGreen, mark=none] table{qasc_te.dat};
    \end{groupplot}
\end{tikzpicture}
\caption{\textbf{The huge performance mismatch between training and validation sets by epochs.}}
\label{figure:mismatch}
\vspace{0.5em}
\centering
\pgfplotsset{
width=0.25\textwidth,
my style/.style={
            every axis plot/.append style={line width=0.8pt},
            font=\tiny,           
            symbolic x coords={10,20,40,60,80,100},
            xtick=data,
            xmin=10,xmax=100,
            ymin=40, ymax=100,
            grid,
            grid style={dotted},},
my legend style/.style={
            font=\scriptsize,
            legend entries={
                Machine Rationales,
                Human Rationales,     
            },
            legend style={
                draw=none,
                at={([yshift=2pt]1.2,1.1)},
                anchor=south,
            },
            legend columns=2,
        },
        every axis title/.append style={at={(0.5,-0.55)}},
        every axis x label/.style={at={(0.3,0.01)},anchor=south west},
        every axis y label/.style={at={(-0.15,0.0)},anchor=south west, rotate=90}
        }
\begin{tikzpicture}
\centering

\begin{groupplot}[
    group style={group size=2 by 1, 
                horizontal sep = 1cm,
                vertical sep = 1.2cm}]
    \nextgroupplot[my style, 
                  my legend style, 
                 y unit={\%},
                 x unit={\%},
                 xlabel=Training Ratio,
                 ylabel=Accuracy, 
                 title={\footnotesize{(a) CSQA}}]
        \addplot[color=BurntOrange, mark=none] table{csqa_cr_m.dat}; 
        \addplot[color=JungleGreen, mark=none] table{csqa_cr_h.dat}; 
    \nextgroupplot[cycle list shift=1, 
                   my style, 
                   title={\footnotesize{(b) ECQA }}]
        \addplot[color=JungleGreen, mark=none] table{ecqa_cr_h.dat}; 
        \addplot[color=BurntOrange, mark=none] table{ecqa_cr_m.dat};
    \end{groupplot}
\end{tikzpicture}
\caption{\textbf{Performance of human or machine rationales by training ratios.} Though we add up the data points, the performance of the reasoning model remains consistent once it meets certain performance thresholds. }
\label{figure:mismatch_training_ratios}
\vspace{-0.5em}
\end{figure}

Language models (LMs) have achieved significant progress on sophisticated reasoning tasks requiring commonsense knowledge or selecting the best answer among tricky multiple choicese~\citep{Wei0SBIXCLZ22, 0010LLWWBCH22}.
Recent advances have enabled LMs to generate explicit free-text rationales and use them to guide task predictions with better search space for candidate answers without injecting additional knowledge~\citep{ZelikmanWMG22, KojimaGRMI22}.

However, it is impossible to consistently obtain perfect rationales from models or even from humans.~\citep{JungQWBB0C22, WangCIC023, ChenB0J0S23}.
Despite the advantage of rationales that can explain model decisions, the human utility of those generated by large-scale LMs is far from satisfactory. \citet{JoshiLRCTNW0023} has recently reported that only 20\% of them are useful for humans to gain additional information to answer questions. Human annotation may enhance the quality of rationales but is often costly yet does not guarantee perfect conditions, either.

Above all, learning the patterns of rationales that are generated based on a tremendous number of different statements is almost improbable. By nature, rationales carry varying normative concepts, corresponding to various kinds of reasoning to justify thinking or action in a certain way~\citep{wedgwood2017value}.
The diversity leads models to face the aleatoric uncertainty (i.e., data uncertainty) that identifies ambiguity or randomness inherent in the observation, causing the difficulty of learning the rationales, as shown in Figure~\ref{figure:mismatch}. The Figure~\ref{figure:mismatch_training_ratios} shows that the performance does not significantly improve with additional data due to the noisiness in the dataset even with human rationales.

Our study aims to investigate how the ambiguous rationales play in model performances of natural language reasoning.
We explore how differently human-annotated and machine-generated rationales impact model reasoning performances. We then introduce a method of simply dealing with unclear machine-generated rationales, proposing a suitable two-system reasoning mechanism depending on the level of ambiguity. Because the quantity of the data does not improve the quality of the data, we target to work with the suboptimal quality of the rationales for faithful enough reasoning rather than improving the quality of the rationales itself. 
\section{Method}

\begin{figure}[t]
    \centering
    \includegraphics[scale=0.49]{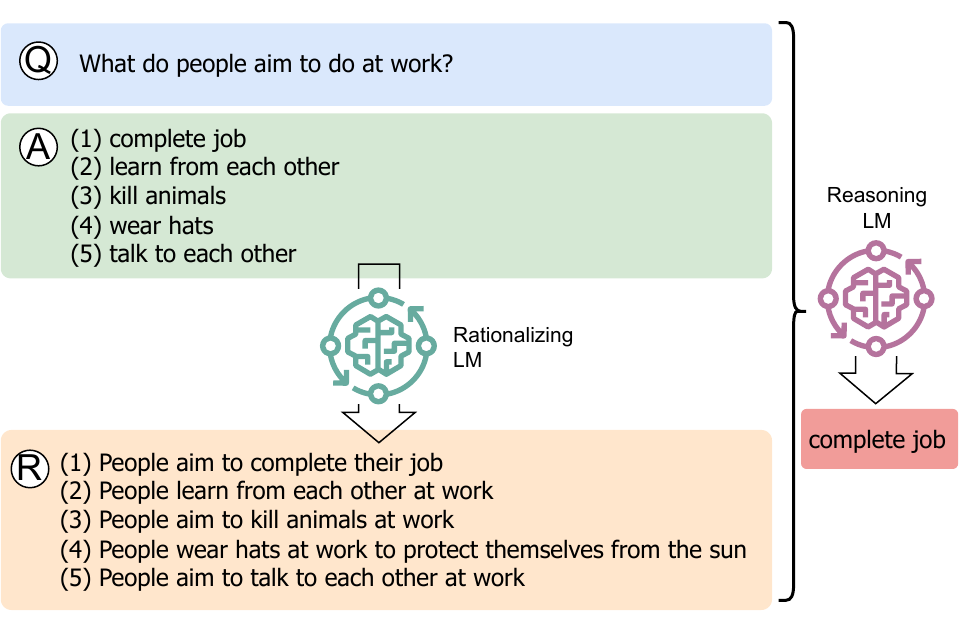}
    \caption{\textbf{An example of reasoning task on generated rationales.} The rationalizing LM generates rationales of the answer choices and the reasoning LM predicts answers on the given question, answer choices, and the generated rationales.}
    \label{fig:aura_example}
\vspace{-1em}
\end{figure}

We study how the model should deal with ambiguous rationales for reasoning tasks.

\subsection{Natural Language Reasoning with Rationales}
Following \citet{ShwartzWBBC20, WiegreffeMS21, WangCIC023}, we explore multiple-choice question-answering tasks in the form of question $q$ and a set of answer choices $A = \{a_i\}$. For such a task, the model predicts an answer $\hat{a} =\argmax_{a_i \in A} \rho(q,a_i)$ and aims to match the predicted answer with a correct answer choice $a^{\ast} \in A$. Note that $\rho(q,a_i)$ is a plausibility score for each $(q, a_i)$ pair. 

Given rationales, the reasoning model learns to output plausibility score $\rho(q, A, a_i, r_i)$ such that a question $q$, the answer choices $A$, answer candidate $a_i\in A$, and rationale $r_i$. 
The reasoning model computes $\rho_i$ by aggregating the probabilities P of tokens $t_i^j$ where $a_i$'s tokens $t_i=[t_i^1, t_i^2, t_i^3,...,t_i^{|a_i|}]$:
\begin{equation}
    \rho_i = \Sigma_{j=1}^{|a_i|}\log P(t_i^j|t_i^{j-1}, ..., t_i^2, t_i^1, q, A, r_i).
\label{eq:normal_probability}
\end{equation}

The aggregated probabilities $P(a_i|q, A, R)$ normalize $p_i$ by using softmax function as $P(a_i|q, A, R)=e^{\rho_i}/\Sigma_{j=1}^{|A|}e^\rho_j$.  The standard training objective is to maximize the likelihood of the correct answer choice using cross-entropy loss, computed as:
\begin{equation}
    \textit{L}=-\Sigma_{a_i \in A}{Q(a_i|q, A)\log P(a_i|q, A, R)}
\end{equation}
where $Q(a_i|q, A)$ is 1 if $a_i=a^*$ and $0$ otherwise.

We mainly explore the reasoning module that produces the ideal outcome given rationales with various qualities.

\subsection{Ambiguous Rationales}
We calculate the entropy values of the rationales, using model prior beliefs as informativeness on multiple choices $A$, and define those with high entropy values as ambiguous.

\paragraph{Entropy.} 
In information theory~\cite{Shannon48}, the entropy measures information or uncertainty inherent to the possible outcome:
\begin{equation}
    H =-\sum_{x} p(x) \log p(x)
\end{equation}
where $p_i$ is the probability of the event occurring and $\log p$ shows the informativeness of the event.

In the rationale entropy estimation, we use the prior beliefs as informativeness, $\log P(a_i|x_i, \tilde{\theta})$, and the posterior beliefs as the probability of the event occurring, $P(a_i|x_i, \hat{\theta}_\text{MLE})$ as illustrated:
\begin{equation}
     H(x) = - P(a_i|x_i, \hat{\theta}_\text{MLE})\log P(a_i|x_i, \tilde{\theta})
     \label{eq:entropy1}
\end{equation}
We aim to measure how the posterior beliefs are uncertain about what prior beliefs have high informativeness.

\paragraph{Model Prior and Posterior Beliefs.} 
LMs have shown that their parameters might contain a significant body of factual and commonsense knowledge as they get trained on a massive volume of text~\citep{Wei0SBIXCLZ22, CohenGBG23}. We call the internal knowledge that the models have had before being fine-tuned with the training set a prior belief of the model $\tilde{\theta}$. After the same model gets trained with a training set, we call their knowledge a posterior belief of the maximum likelihood model $\hat{\theta}_{\text{MLE}}$. 
The prior and posterior belief probabilities come from the model plausibility score as in Equation~\ref{eq:normal_probability}. We normalize the probability scores over the answer choices for each question.
We do not particularly measure model prior and posterior beliefs in the context of Bayesian settings but rather a deterministic approach with a single model.


\paragraph{Rationale Ambiguity.} 
We use the expected value of entropies with the entire data points as a threshold to determine ambiguous and unambiguous rationales within the distribution of how data points interact with model training.
\begin{equation}
    R(x) = 
    \begin{cases}
        r_{\text{unambiguous}}, \hspace{1em}\text{ if } h(x_i) < \tau\\
        r_{\text{ambiguous}}, \hspace{1em}\text{ if } h(x_i) \geq \tau
    \end{cases}
    \label{eq:amb_unamb}
\end{equation}
where $\tau$ is $\mathcal{E}[H(x)]$. The randomness and ambiguity mainly come from their indecisive behaviors between prior and posterior beliefs, comparing before and after the model learning.

\subsection{AURA: Reasoning with Ambiugous Rationales.} 

We utilize a two-stage reasoning system depending on the ambiguity scores of rationales from Equation~\ref{eq:amb_unamb}. We first use a pre-trained model to train a reasoning module on the entire dataset  (Reasoning 1). We use the pre-trained model $\tilde{\theta}$ as prior beliefs and the trained model $\hat{\theta}_{\text{MLE}}$ as posterior beliefs to calculate rationale entropy. We then train the same pre-trained model again on the selected ambiguous rationales to learn unlearned or underlearned data points that have high entropy scores (reasoning 2). It works as ensemble learning as illustrated in Equation~\ref{ensemble}.
\begin{equation}
\begin{aligned}
    \{P(y|x^\ast, \theta^{(t)}\}_{t=1}^{T=2} \rightarrow 
    \{P(y|W^{(t)})\}_{t=1}^{T=2}, \\
    \newline
    W^{(t)} \sim P(W|x^t,D)
\end{aligned}
\label{ensemble}
\end{equation}
where $t=1$ and $t=2$ indicate Reasoning 1 and 2 procedures, respectively. 

\section{Experiments}
In this section, we observe the superiority of AURA which deals with aleatoric uncertainty in rationales for reasoning tasks.

\subsection{Experimental Setup}
We use five different datasets and seven types of baseline approaches to explore AURA. We conducted the experiments with 5 other random seeds and reported the average scores for the results. 

\paragraph{Datasets}

We experiment with four commonsense question-answering datasets that require machine rationales in addition to one benchmark that includes human rationales: (a) CSQA~\citep{TalmorHLB19} a 5-choice dataset that requires knowledge extracted from ConceptNet. (a-1) ECQA~\citep{aggarwal2021explanations} explains each choice to questions from CSQA but with human rationales.
(b) StrategyQA~\citep{GevaKSKRB21} is a dataset that requires an advanced level of reasoning strategies to produce binary (yes/no) answers.
(c) OpenbookQA~\citep{MihaylovCKS18} is a 4-choice dataset to test model ability as a form of open book testing.
(d) QASC~\citep{KhotCGJS20} is a complicated 8-choice question set.

\paragraph{Baselines} We consider four rationalize-then-reason pipeline approaches similar to AURA as baselines. We also compare our approach with those with self-rationales or without any rationales. (1) Without rationales: fine-tune T5-base model without providing rationales. 
(2) Answer upon self-rationales using a medium-sized LM: make GPT-neox (20B)~\citep{black-etal-2022-gpt} learn from a few examples to generate machine rationales and the following answers. 
(3) Answer upon self-rationales using a large-scale LM: same mechanism as (2) but use GPT3 (175B)~\citet{Wei0SBIXCLZ22}.
(4) NILE~\citep{KumarT20}: a method of language inference on label-specific explanations. Following the standard setup, we fine-tune NILE with a T5 (3B) model, unlike other baselines that use a T5-base model. 
(5) Train on prompted rationales with a standard training method: use GPT-neox but obtain answers not from the same model but from a separate reasoning model, T5-base in our experiment.
(6) Dropout context is the same as (5) but randomly drops out from the input question while training the reasoning model~\citep{WangCIC023}.
(7) same as (5) but trained with counterfactual regularization objective.
We report the results presented by \citet{WangCIC023}.

\paragraph{Implementation Details}
For machine-generated rationales, we follow the same setup with \citet{WangCIC023} for a fair comparison, 7 manually-annotated prompt examples to GPT-neox (20B) for each dataset. We adopt the T5-base model for the reasoning module.

\subsection{Evaluation}

\begin{table*}[t!]
\footnotesize
\begin{center}
  \begin{tabular}{ p{6cm}  c  c  c  c  c}
    \toprule
              & \multicolumn{4}{c}{ Dataset \vspace{0.5em} }& \\ 
                \hspace{5em} Method & CSQA & StrategyQA& OBQA & QASC & Average \\ \midrule
             \textbf{\footnotesize w/o Rationales}  ... (1)   &  58.68 & 58.12 & 55.85 & 35.58 & 52.06 \\ \midrule[0.01pt] 
            \textbf{\footnotesize Self-Rationalization} \\
            {\hspace{2em}\footnotesize  GPT-neox, 20B} ... (2)                 & 38.41 & 55.31 & 33.80 & 32.61 & 45.89 \\ 
            {\hspace{2em}\footnotesize  GPT3, 175B (Chain-of-Thought)} ...(3)  & 63.50 & 64.40 & - & - & -\\ \midrule[0.01pt] 
            \textbf{\footnotesize Pipeline Rationalization} \\
            {\hspace{2em}\footnotesize NILE} ...(4)                 & 57.60 & 57.31 & - & - & - \\ 
            {\hspace{2em}\footnotesize Standard Training} ...(5)               & 59.48 & 57.11 & 56.65 & 37.50 & 52.69\\ 
            {\hspace{2em}\footnotesize Dropout Context} ...(6)                   & 59.64& 51.45& 57.55 & 35.37 & 51.00 \\ 
            \hspace{2em}{\footnotesize PINTO}\hspace{.2em}{\scriptsize Counterfactual Regularization}\dag ...(7)    & 61.67 & 60.87 & 58.85 & 37.82 & 54.80 \\ 
            \hspace{2em}{\footnotesize AURA}\ddag &\textbf{67.13 }{\scriptsize(+5.46)} & \textbf{76.35 }{\scriptsize(+15.48)} & \textbf{65.89 }{\scriptsize(+7.04)}&  \textbf{40.82 }{\scriptsize(+3.00)} & \textbf{62.55 }{\scriptsize(+7.04)}\\ 
    \bottomrule
\end{tabular}
\vspace{-0.5em}
\caption{\textbf{Comparison of task performances reported with accuracy.} We bold the best results and report the performance gain by AURA from the best results among existing pipeline rationalization methods with (+ sign). \dag: a current state-of-the-art method in the pipeline rationalization, \ddag: our proposed method.}
\label{table:main}
\end{center}
\vspace{-1em}
\newcommand{\veryshortarrow}[1][5pt]{\mathrel{%
   \vcenter{\hbox{\rule[-.2pt]{#1}{0.5pt}}}%
   \mkern-4mu\hbox{\usefont{U}{lasy}{m}{n}\symbol{41}}}}
\footnotesize
\begin{center}
  \begin{tabular}{ p{5.5cm}  c c c c c c }
    \toprule
             & \multicolumn{5}{c}{ Dataset (\textit{source $\rightarrow$ target}) \vspace{0.5em} } \\ 
             &\hspace{-0.8em}CSQA$\veryshortarrow$ &\hspace{-0.8em}CSQA$\veryshortarrow$ & \hspace{-0.8em}OBQA$\veryshortarrow$ &\hspace{-0.8em}QASC$\veryshortarrow$ & \hspace{-0.8em}QASC$\veryshortarrow$\\
             \hspace{5em}Method                    &\hspace{0.8em}OBQA  &\hspace{0.8em}QASC &\hspace{0.3em}CSQA &\hspace{0.8em}CSQA &\hspace{0.8em}OBQA \\ \midrule
            \textbf{w/o Rationales}  ... (1)                & 32.05 &  39.17 & 24.87 & 45.74 & 34.90 \\ \midrule[0.01pt] 
            \textbf{\footnotesize Pipeline Rationalization} \\ 
            \hspace{2em}NILE ... (4)                  & 32.40  & 40.93 & - & - & - \\ 
            \hspace{2em}Standard Training.... (5)  & 31.05 & 40.04 & 25.37 & 47.71 & 34.50 \\ 
            \hspace{2em}Dropout Context ... (6)                   & 32.30 & 38.85 & 23.01 & 44.27 &32.90\\  
            \hspace{2em}{\footnotesize PINTO}\hspace{.2em}{\scriptsize Counterfactual Regularization}\dag ... (7)    & 34.90 & 42.25 & 27.66 & 48.03 & 35.75  \\  
            \hspace{2em}{\footnotesize AURA}\ddag & \textbf{49.80 }{\scriptsize(+14.90)} & \textbf{43.20 }{\scriptsize(+0.95)} &\textbf{42.91 }{\scriptsize(+15.25)} & \textbf{52.48 }{\scriptsize(+4.45)} & \textbf{51.40 }{\scriptsize(+15.65)}\\
    \bottomrule
\end{tabular}
\vspace{-0.5em}
\caption{\textbf{OoD Test.} Performance on a target dataset after training on a source dataset, denoted as \textit{source $\rightarrow$ target}.}
\label{table:ood}
\end{center}
\vspace{-1.5em}
\end{table*}

We test AURA on In-Distribution (ID) and Out-of-Distribution (OoD) settings and different training ratios to investigate the superiority of AURA. 

\paragraph{Comparison against Other Baseline Methods.} 

We first evaluate AURA in ID settings where we use the target dataset for rationalizing and reasoning modules and test the performance on unseen examples of the same dataset. We have observed results as shown in Table~\ref{table:main}. AURA achieves strong task performance compared to other baseline methods on almost all four benchmark datasets. For CSQA and StrategyQA, AURA outperforms the self-rationalizing method with chain-of-thought (GPT3, 175B) by 3.63\% and 11.95\% accuracy points, respectively. AURA outperforms the state-of-the-art approach with the counterfactual regularization method on four out of four benchmarks and achieved the new state-of-the-art by increasing the accuracy score by 7.75 \% points on average.

\paragraph{Out-of-distribution (OoD) Performance}
We now evaluate AURA in OoD settings where we train on source datasets and test on different target datasets to investigate how AURA helps models generalize well to reason on rationales rather than memorize the concepts or answers. 
Table~\ref{table:ood} shows that AURA outperforms all existing methods on the entire benchmark datasets by an average score of 10.24\% and at most 15.65\% of accuracy.
Following previous work, we adopt the intuition that the models produce better OoD results than others if they utilize rationales faithfully. 

\paragraph{Training Ratios.}
\usepgfplotslibrary{units}
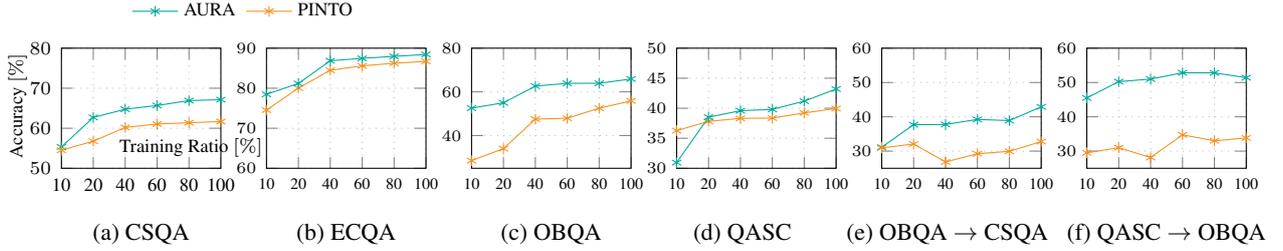
\begin{figure*}[t!]
\centering
\hspace{-1em}
\pgfplotsset{
width=0.23\textwidth,
my style/.style={
            font=\tiny,           
            symbolic x coords={10,20,40,60,80,100},
            xtick=data,
            xmin=10,xmax=100,
            grid,
            grid style={dotted},},
my legend style/.style={
            font=\scriptsize,
            legend entries={
                {AURA \hspace{2em}},    
                PINTO
            },
            legend style={
                draw=none,
                at={([yshift=2pt]1.15,1.1)},
                anchor=south,
            },
            legend columns=7,
        },
        every axis title/.append style={at={(0.5, -0.85)}},
        every axis x label/.style={at={(0.3,0.01)},anchor=south west},
        every axis y label/.style={at={(-0.15,0.0)},anchor=south west, rotate=90}
        }
\begin{tikzpicture}
\centering

\hspace{-1em}
\begin{groupplot}[
    group style={group size=6 by 1, 
                horizontal sep = 0.6cm,
                vertical sep = 1cm}]
    \nextgroupplot[my style, 
                  my legend style, 
                  x unit={\%},
                 y unit={\%},
                 xlabel=Training Ratio,
                 ylabel=Accuracy, 
                 ymin=50, ymax=80,
                 title={\footnotesize{(a) CSQA}}]
        \addplot[color=JungleGreen, mark=asterisk] table{csqa_aura_m.dat}; 
        \addplot[color=BurntOrange, mark=asterisk] table{csqa_cr_m.dat};   
    \nextgroupplot[cycle list shift=1, 
                    my style, 
                     ymin=60, ymax=90,
                   title={\footnotesize{(b) ECQA }}]
        \addplot[color=BurntOrange, mark=asterisk] table{ecqa_cr_h.dat}; 
        \addplot[color=JungleGreen, mark=asterisk] table{ecqa_aura_h.dat};
        \nextgroupplot[cycle list shift=2, 
                   my style, 
                    ymin=25, ymax=80,
                   title={\footnotesize{(c) OBQA}}]   
        \addplot[color=JungleGreen, mark=asterisk] table{obqa_aura.dat}; 
        \addplot[color=BurntOrange, mark=asterisk] table{obqa_cr.dat}; 
        \nextgroupplot[cycle list shift=3, 
                   my style, 
                    ymin=30, ymax=50,
                   title={\hspace{-2em} \footnotesize{(d) QASC}}]
        \addplot[color=BurntOrange, mark=asterisk] table{qasc_cr_m.dat};   
        \addplot[color=JungleGreen, mark=asterisk] table{qasc_aura_m.dat};
        \nextgroupplot[cycle list shift=2, 
                   my style, 
                     ymin=25, ymax=60,
                   title={\hspace{-2em} \footnotesize{(e) OBQA $\rightarrow$ CSQA}}]
 \addplot[color=JungleGreen, mark=asterisk] table{obqa_csqa_aura.dat}; 
        \addplot[color=BurntOrange, mark=asterisk] table{obqa_csqa_cr.dat};
        \nextgroupplot[cycle list shift=3, 
                   my style, 
                     ymin=25, ymax=60,
                   title={\footnotesize{(f) QASC  $\rightarrow$ OBQA}}]
         \addplot[color=JungleGreen, mark=asterisk] table{qasc_obqa_aura.dat}; 
        \addplot[color=BurntOrange, mark=asterisk] table{qasc_obqa_cr.dat}; 
    \end{groupplot}
\end{tikzpicture}
\caption{\textbf{Performance changes depending on different training ratios.} The performance of in-distribution settings with (a) -- (d) and of out-of-distribution settings with (e) -- (f).}
\label{figure:ratios}
\vspace{-1em}
\end{figure*}

We test AURA on different training ratios to see its performance in low-resource settings. We observe that AURA is strong on limited resource settings on all four benchmarks as shown in Figure~\ref{figure:ratios}. We observe AURA outperforms PINTO, the state-of-the-art rationalize-then-reason method, in both ID and OoD settings. Due to the nature of aleatoric uncertainty in rationales, the performance gains by AURA are pretty consistent over different training ratios.

\subsection{Analysis}

\begin{figure}[t!]
\hspace{-2em}
   \begin{tikzpicture}
   \begin{axis}[
        ybar,
        bar width=.7cm,
        width=0.55\textwidth,
        height=0.23\textwidth,
        legend style={at={(0.5,1.3), font=\scriptsize},
            anchor=north,legend columns=-1},
        symbolic x coords={a,b,c,d},
        xtick=data,
        nodes near coords,
        xticklabel=\empty,
        every node near coord/.append style={font=\scriptsize, color=black},
        ymin=55,ymax=100,
        ylabel={Accuracy\%},
        xlabel={Test: Human Rationales (Top) and Machine Rationales (Bottom)},
        enlarge x limits=0.5,
        ylabel style = {font=\footnotesize},
        xlabel style = {font=\scriptsize},
        xticklabel style=\empty,
        yticklabel style = {font=\scriptsize},
        bar shift=5pt,
    ]
    \addplot[BurntOrange, fill=BurntOrange!50!, ] coordinates{(a, 59.48)};
    \addplot[brown, fill=brown!70!, ] coordinates{(b,64.01)};
    \addplot[JungleGreen!100!, fill=JungleGreen!70!, ] coordinates{(c,67.13)};
    \addplot[NavyBlue!100!, fill=NavyBlue!80!, ] coordinates{(d,69.47)};
    \legend{Machine, Human, AURA+M, AURA+H}
    \end{axis}
    
    \begin{axis}[
        hide axis, 
        ybar,
        bar width=.7cm,
        width=0.55\textwidth,
        height=0.23\textwidth,
        legend style={at={(0.5,1.3), font=\scriptsize},
            anchor=north,legend columns=-1},
        symbolic x coords={a,b,c,d},
        xtick=data,
        nodes near coords,
        xticklabel=\empty,
        every node near coord/.append style={font=\scriptsize, color=black},
        ymin=55,ymax=100,
        ylabel={Accuracy\%},
        xlabel={Test: Human Rationales (Top) and Machine Rationales (Bottom)},
        enlarge x limits=0.5,
        ylabel style = {font=\footnotesize},
        xlabel style = {font=\scriptsize},
        xticklabel style=\empty,
        yticklabel style = {font=\scriptsize},
        bar shift=5pt,
    ]
    \addplot[BurntOrange, fill=BurntOrange!50!, pattern=north west lines, pattern color=.] coordinates{(a, 81.92)};
    \addplot[brown, fill=brown!70!, pattern=north west lines, pattern color=.] coordinates{(b,86.21)};
    \addplot[JungleGreen!100!, fill=JungleGreen!70!, pattern=north west lines, pattern color=.] coordinates{(c,86.81)};
    \addplot[NavyBlue!100!, fill=NavyBlue!80!, pattern=north west lines, pattern color=.] coordinates{(d,88.42)};
    \end{axis}
\end{tikzpicture}
\vspace{-2em}
\caption{\textbf{Performance gains by AURA depending on training and testing rationales.} M: machine-generated rationales, H: human-written rationales. The legend shows types of training rationales with or without AURA. Without AURA is standard training.}
\label{fig:machine_human}
\vspace{1em}
\hspace{-2em}
\pgfplotstableread[row sep=\\,col sep=&]{
    test & A & B  & C & D\\
    Test: OBQA    & 34.91  & 32.20 & 49.80 & 49.85  \\
    Test: QASC     & 42.25  & 34.83 & 43.20 & 35.56 \\
    }\mydata
\begin{tikzpicture}
    \begin{axis}[
            ybar,
            bar width=.7cm,
            width=0.55\textwidth,
            height=0.23\textwidth,
            legend style={at={(0.5,1.3), font=\scriptsize},
                anchor=north,legend columns=-1},
            symbolic x coords={Test: OBQA,Test: QASC},
            xtick=data,
            nodes near coords,
            every node near coord/.append style={font=\scriptsize, color=black},
            ymin=30,ymax=55,
            ylabel={Accuracy\%},
            xlabel={Reasoning Method + Training Rationales},
            enlarge x limits=0.5,
            ylabel style = {font=\footnotesize},
            xlabel style = {font=\scriptsize},
            xticklabel style={font=\scriptsize},
            yticklabel style = {font=\scriptsize},
        ]
        \addplot[BurntOrange, fill=BurntOrange!50!, pattern=north west lines, pattern color=.] table[x=test,y=A]{\mydata};
        \addplot[brown, fill=brown!70!, pattern=north east lines, pattern color=.] table[x=test,y=B]{\mydata};
        \addplot[JungleGreen!100!, fill=JungleGreen!70!, pattern=north west lines, pattern color=.] table[x=test,y=C]{\mydata};
        \addplot[NavyBlue!100!, fill=NavyBlue!80!, pattern=north east lines, pattern color=.] table[x=test,y=D]{\mydata};
        \legend{PINTO+M, PINTO+H, AURA+M, AURA+H}
    \end{axis}
\end{tikzpicture}
\vspace{-0.7em}
\caption{\textbf{Performance comparison between human and machine rationales in OoD settings.} For the competitive comparison, we used PINTO, the state-of-the-art method, to compare with AURA as the reasoning method in this experiment.}
\label{fig:machine_human_ood}
\vspace{-1em}
\end{figure}

\paragraph{RQ1. Does the quality rationales contribute more than the good reasoning mechanisms to the overall performance? } Robust reasoning method matters more than obtaining quality rationales for predicting right answers. As shown in Figure~\ref{fig:machine_human}, the performance gain by AURA (+7.65\%) is greater than that of better rationales, using human rationales instead of machine rationales (+4.53\%) when tested on machine rationales. Similarly, when tested on human rationales, AURA (+4.89\%) works slightly better than better rationales (+4.29\%).

In OoD settings, the statement above is more solid than in ID settings as shown in Figure~\ref{fig:machine_human_ood}. Superior rationales in ID settings do not guarantee satisfaction in OoD settings (34.91\%$\rightarrow$32.2\%, 42.25\%$\rightarrow$34.82\%). On the other hand, the performance gains by better rationales (-2.71\%, -7.42\%) underperform those by AURA (+14.89\%, +0.95\%). It is worth noting that AURA always guarantees slight performance gains without any performance degradation. This shows that better reasoning is capable of covering unsatisfactory rationalization.

\paragraph{RQ2. Do quality rationales provide a better influence on training or inference?} We have observed in Figure~\ref{fig:machine_human} that the quality of rationales does not improve the model learnability much compared to what it does to model inference ability. Standard training on human rationales has increased model performance by 4.53\% or 4.29\% from machine rationales. Both models trained on machine and human rationales have improved their performance accuracy by 22.44\% and 22.2\% each when tested on human rationales instead of machine rationales.  

If we have a limited amount of quality rationales, we better use them while inferring the answers to obtain the most decent performance level among other methods using them. However, in realistic scenarios, we cannot tell if the seemingly ideal rationales are still satisfactory to the question-answer pairs that we want the answers to. As discussed in RQ1, if the inferring data are OoD, the results would be disappointing.

Hence our research statement is that it is very important to study how we deal with ambiguous, confusing rationales for natural language reasoning. AURA indeed is strong on both OoD settings and unsatisfactory rationales.

\section{Conclusion}
Generating impeccable rationales is often impossible, and learning the patterns of rationales is almost improbable. 
We first explore how ambiguous it is for language models to learn rationales for reasoning tasks, both with human-annotated and machine-generated rationales. We then introduce a method of simply dealing with unclear rationales, proposing a suitable two-system reasoning mechanism depending on the level of ambiguity. We empirically argue that AURA produces robust performance superiority against the adversarial quality of rationales and low-resource settings.

\section{Limitations}
Our approach has been tested mainly on medium-sized LMs due to computational cost. We have also focused on commonsense reasoning tasks rather than those requiring domain-specific knowledge. This is due to our approach of exploring the benefits of model prior knowledge from pre-trained LMs. 

\section*{Acknowledgements} 
This research was supported and funded by Classting Inc. The author thanks the members of the Classting AI Research, Jaesang Yoo, Yohaan Yoon, Cheoleui Hong, and Heeseok Jung, for their support for this work in various ways.

\bibliography{coling}

\end{document}